\documentclass[10pt,twocolumn,letterpaper]{article}

\usepackage[algorithms]{wacv}

\usepackage{tabularx}
\newcolumntype{C}{>{\centering\arraybackslash}X}

\usepackage{amsmath,amssymb,amsfonts}
\usepackage{mathtools}
\usepackage{bm}
\usepackage{booktabs}
\usepackage{multirow}
\usepackage{makecell}
\usepackage{colortbl}
\usepackage{xcolor}
\usepackage{tikz}
\usetikzlibrary{arrows.meta,positioning,calc,backgrounds,fit,decorations.pathmorphing,shapes.geometric,patterns}
\usepackage{graphicx}
\usepackage{amsthm}

\newtheorem{theorem}{Theorem}
\newtheorem{proposition}{Proposition}
\newtheorem{corollary}{Corollary}
\newtheorem{definition}{Definition}

\definecolor{hlbest}{RGB}{222,235,247}
\definecolor{hlsecond}{RGB}{240,240,240}
\definecolor{hlours}{RGB}{255,244,222}
\definecolor{headblue}{RGB}{31,78,121}
\definecolor{accent}{RGB}{192,0,0}
\definecolor{deepgreen}{RGB}{0,110,70}
\newcommand{\best}[1]{\textbf{#1}}
\newcommand{\sand}[1]{\underline{#1}}
\newcommand{\up}{\textcolor{headblue}{$\uparrow$}}
\newcommand{\down}{\textcolor{headblue}{$\downarrow$}}
\newcommand{\dimp}[1]{\textcolor{accent}{\scriptsize\,(#1)}}

\newcommand{\Dv}{\Delta\bm{v}}
\newcommand{\sg}{\mathrm{sg}}
\newcommand{\agave}{\textsc{AGAVE}}
\newcommand{\iicfg}{I\textsuperscript{2}\text{-}CFG}

\definecolor{wacvblue}{rgb}{0.21,0.49,0.74}
\usepackage[pagebackref,breaklinks,colorlinks,allcolors=wacvblue]{hyperref}

\def\wacvPaperID{2223}
\def\confName{WACV}
\def\confYear{2027}

\title{Guidance Is a Field, Not a Scalar:\\
Interval-Integrated Classifier-Free Guidance for One-Step Video Generation}

\author{Zijie Meng \qquad Xixin Cao\\
Peking University\\{\tt\small ymlf@stu.pku.edu.cn \qquad cxx@ss.pku.edu.cn}}

\begin{document}
\maketitle

\begin{abstract}
\textit{One-step generators built on MeanFlow synthesize a video in a single function evaluation, yet inherit a rigid treatment of classifier-free guidance (CFG): the guidance scale is frozen during training or, at best, exposed as a single scalar. This conflicts with mounting evidence that the optimal guidance is not a constant but a \emph{field} that varies with the noise level, with spatial content, and, for video, with frame position. Every existing remedy presupposes a multi-step sampler along which guidance can be re-scheduled---an assumption one-step models violate by construction. We resolve this tension through a single observation: because the MeanFlow target is an integral over a noise interval, an arbitrary guidance field can be absorbed into the target itself and compiled into one evaluation. We formalize this as the \emph{interval-integrated CFG} identity, prove that the compiled guidance carries no schedule-discretization error and can thus match or exceed any finite multi-step sampler, and instantiate it as \agave{}, which equips the average-velocity field with a learnable temporal schedule, a free motion-saliency spatial modulation, and a frame-position asymmetry. Adding fewer than ten scalars at zero inference cost, \agave{} surpasses all one-step baselines on VBench, UCF-101 and MSR-VTT while matching multi-step guided sampling. Project repository: \url{https://github.com/YMlinfeng/AGAVE}.}
\end{abstract}

\section{Introduction}
\label{sec:intro}

Diffusion and flow-matching models have become the backbone of modern visual generation~\cite{ho2020ddpm,song2021sde,lipman2023flow,liu2023rectified,rombach2022ldm,esser2024sd3,peebles2023dit}, and their extension to video has produced systems of remarkable fidelity~\cite{ho2022vdm,blattmann2023svd,brooks2024sora,yang2024cogvideox,kong2024hunyuanvideo,meng2026argus,meng2026omnidrive,wan2025wan}. Camera control has also advanced through scale-calibrated transfer~\cite{meng2026parascale}, multi-shot camera cloning in OmniDirector~\cite{liu2026omnidirector}, and video re-shooting in TARS~\cite{liu2026tars}. Their central weakness is sampling cost, since faithfully integrating the generative ordinary differential equation still requires many network evaluations. A long line of distillation and consistency methods has attacked this bottleneck~\cite{salimans2022progressive,song2023consistency,songdhariwal2024ict,lu2025scm,frans2025shortcut,yin2024dmd,yin2024dmd2}, and most recently MeanFlow~\cite{geng2025meanflow} reframed one-step generation as the regression of an \emph{interval-averaged} velocity rather than an instantaneous one, achieving an FID of $3.43$ with a single function evaluation on ImageNet $256^2$. Its successor iMF~\cite{geng2025imf} stabilized the objective and pushed the one-step FID to $1.72$, while NVIDIA's rCM~\cite{zheng2026rcm} scaled the same continuous-time machinery to ten-billion-parameter video diffusion, distilling Wan2.1~\cite{wan2025wan} into a few-step generator that rivals its multi-step teacher on VBench~\cite{huang2024vbench}.

This progress exposes a problem that has been largely overlooked. Classifier-free guidance~\cite{ho2022cfg, meng2026decoupling} is indispensable for conditional fidelity, but it interacts awkwardly with one-step models. The original MeanFlow bakes a \emph{fixed} guidance scale into its target, trading away the ability to retune the single most important inference knob~\cite{geng2025meanflow}. iMF improves on this by treating the scale as a conditioning variable, which restores test-time flexibility and confirms that the best scale is itself a moving target, drifting with model size, training length and step count~\cite{geng2025imf}. Yet even iMF treats guidance as a \emph{scalar}, applied uniformly across the entire generative interval.

A scalar is the wrong object. Analytical and empirical studies show that strong guidance early erodes diversity while strong guidance late suppresses fine-grained variation~\cite{jin2025stagewise}, motivating a \emph{time-varying} schedule. Concrete realizations abound: restricting guidance to a limited interval improves sample quality~\cite{kynkaanniemi2024interval}, Beta-shaped schedules that peak in mid-denoising improve FID at fixed alignment~\cite{malarz2025betacfg}, nonlinear power-law guidance improves robustness and diversity~\cite{pavasovic2025powerlaw}, CFG-Zero\(^\star\) zeroes the first ODE steps where the flow is still underfitted~\cite{fan2025cfgzero}, projected and adaptive variants remove the over-saturation of high scales~\cite{sadat2025apg}, adaptive guidance can omit redundant unconditional evaluations~\cite{castillo2025adaptive}, and online feedback can select per-step scales~\cite{papalampidi2025dynamic}. For video the field is not even one-dimensional: SV4D~2.0 finds it necessary to \emph{linearly increase} the guidance scale along both the view and the frame axes~\cite{yao2025sv4d2}, and few-step video distillers are notorious for the over-saturation and temporal collapse that a mis-calibrated, uniform guidance induces~\cite{zheng2026rcm,advdistill2026}.

The difficulty is structural. The time-varying schedules above are deployed along a multi-step trajectory. A one-step model has no such trajectory; iMF's scalar conditioning, and even its on/off interval gating, cannot express a guidance that varies \emph{within} the single jump from noise to data. The community thus faces an apparent dichotomy: keep the one-step efficiency and accept a crude scalar guidance, or recover adaptive guidance and pay for many steps.

We show the dichotomy is false, and that its resolution has been hiding in plain sight inside the MeanFlow identity. The decisive move is to stop asking \emph{what number} the guidance scale should be and to ask instead \emph{what function} it should be along the generative trajectory---to treat guidance as a field rather than a scalar. Once guidance is a field, the integral structure of MeanFlow stops being a mere efficiency device and reveals itself as a computational substrate. Because the average velocity \(\bm{u}(\bm{z}_t,r,t)=\tfrac{1}{t-r}\int_r^t \bm{v}(\bm{z}_\tau,\tau)\,d\tau\) is an \emph{integral} over a noise interval, it can absorb an integrand that varies arbitrarily along \(\tau\): replacing the constant-guided instantaneous velocity by one guided with a continuous field \(\omega(\cdot)\) yields a new ground-truth average that still satisfies a MeanFlow-type identity, with the boundary term automatically reading the \emph{local} guidance value. Training a single network to match this target therefore \emph{compiles} an entire time-, space-, and frame-varying schedule into one function evaluation.

This reframing carries a consequence we believe is conceptually deeper than the operator itself. What a multi-step sampler accomplishes by \emph{unrolling} a schedule step by step at inference, a one-step model can accomplish by \emph{integrating} that schedule in closed form once, at training time---an amortization that trades repeated runtime computation for a single integral. And because the integral is exact, the compiled guidance inherits none of the discretization error that every finite scheduler must pay: the one-step model does not merely approximate the multi-step schedule but can provably surpass it. Guidance is only the first and most consequential instance of a broader principle---that any inference-time control expressible along the generative trajectory can be folded into the average-velocity field and recovered for free at deployment. Tellingly, this is something single-time models such as consistency models cannot do, precisely because they expose no interval \([r,t]\) over which a within-trajectory schedule could be integrated; the two-time structure of MeanFlow is the enabling ingredient. We name the operator interval-integrated CFG (\iicfg) and the full method \agave{} (Adaptive Guidance-Aware Velocity Ensembling).

In short, we turn the defining limitation of one-step models---the absence of a trajectory to schedule over---into their decisive advantage, with four contributions:
\begin{itemize}\itemsep3pt
\item \textbf{A new object.} We recast classifier-free guidance for one-step generation from a scalar into a continuous \emph{guidance field} over noise-time, space, and video-frame.
\item \textbf{A new identity.} We prove the \emph{interval-integrated CFG identity}: any such field is losslessly absorbed into the MeanFlow target and compiled into one evaluation, provably free of schedule-discretization error.
\item \textbf{A new method.} We instantiate it as \agave{}---fewer than ten extra scalars, zero inference overhead---which provably subsumes fixed-scale MeanFlow CFG, iMF, and one-step CFG-Zero\(^\star\).
\item \textbf{A SOTA result.} \agave{} is the first one-step generator to rival multi-step adaptive guidance, setting a new state of the art on VBench, UCF-101, and MSR-VTT.
\end{itemize}

\begin{figure*}[t]
\centering
\includegraphics[width=\linewidth]{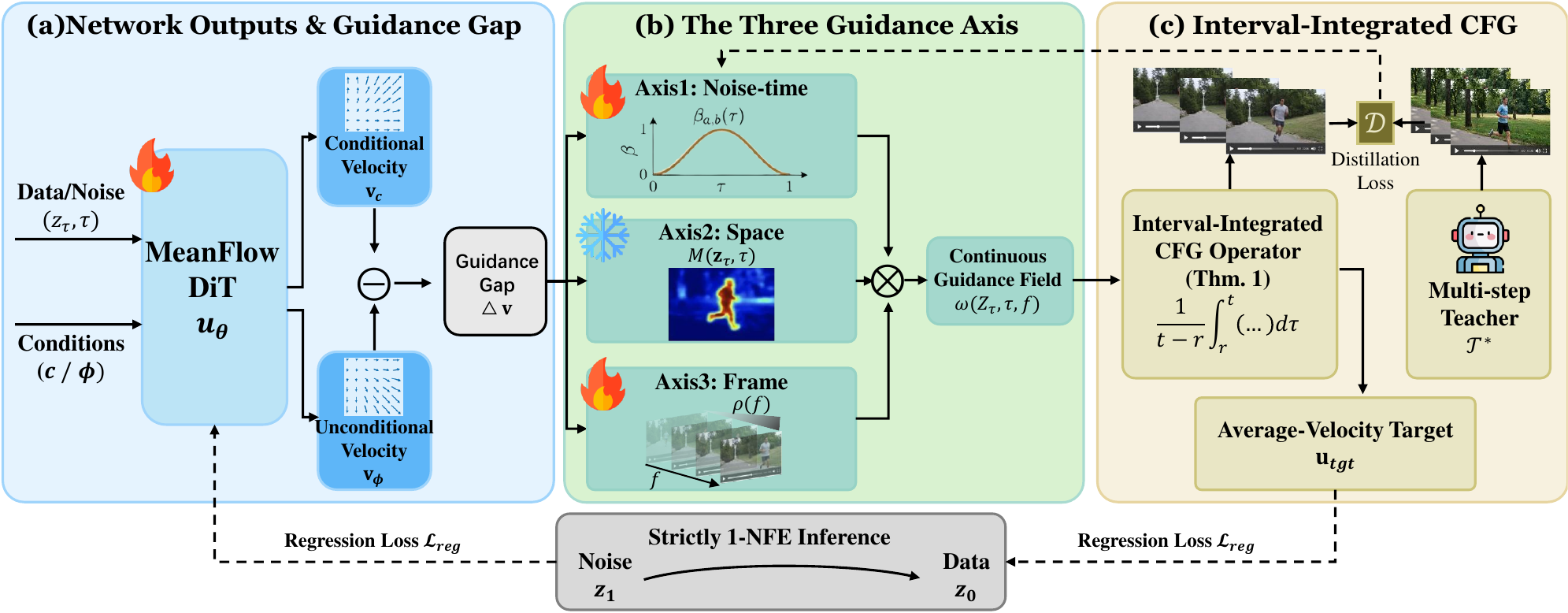}
\vspace{-0.5cm}
\caption{\textbf{Overview of \agave{}.} A single MeanFlow DiT $\bm u_\theta$ emits a conditional/unconditional velocity pair whose guidance gap $\Dv$ is modulated along three axes -- a learnable Beta schedule over noise level $\tau$, a free motion-saliency map $M$ over space, and a learnable frame-position profile $\rho(f)$ -- forming a continuous \emph{guidance field} $\omega(\bm z_\tau,\tau,f)$. The interval-integrated CFG operator (Thm.~\ref{thm:main}) absorbs this field into the average-velocity target, compiling a full time-, space- and frame-varying schedule into one evaluation. The schedule parameters $\phi$ are learned by distilling a multi-step adaptively guided teacher; inference remains strictly 1-NFE.}
\label{fig:hero}
\vspace{-0.5cm}
\end{figure*}

\section{Related Work}
\label{sec:related}

\subsection{One-step and few-step generative modeling}
Reducing the evaluation budget of diffusion and flow models has been pursued through progressive distillation~\cite{salimans2022progressive}, distribution matching~\cite{yin2024dmd,yin2024dmd2}, consistency training~\cite{song2023consistency,songdhariwal2024ict,lu2025scm}, shortcut models~\cite{frans2025shortcut}, and, most relevant here, MeanFlow~\cite{geng2025meanflow}, which learns a long-range average velocity through a differential identity and so removes the need to integrate at inference. iMF~\cite{geng2025imf} recasts the objective as a stable velocity regression and exposes the guidance scale as a conditioning variable; AlphaFlow~\cite{zhang2025alphaflow} shows the MeanFlow objective decomposes into trajectory flow matching and trajectory consistency, whose gradients conflict, and rCM~\cite{zheng2026rcm} scales the JVP-based formulation to billion-parameter video diffusion with a forward-reverse divergence regularizer. Interactive generation further motivates efficient control, from Make a Game~\cite{meng2026make} to LiveAct's multi-agent action interfaces and causal streaming generator~\cite{wang2026liveact}. These one-step guidance formulations either freeze guidance or keep it scalar; none treats guidance as a quantity that varies inside the single jump, which is the gap we close.

\subsection{Adaptive classifier-free guidance}
Beyond the constant scale of vanilla CFG~\cite{ho2022cfg}, stage-wise analysis motivates varying guidance strength along the sampling trajectory~\cite{jin2025stagewise}. This motivates limited-interval guidance~\cite{kynkaanniemi2024interval}, Beta-shaped schedules~\cite{malarz2025betacfg}, nonlinear power-law guidance~\cite{pavasovic2025powerlaw}, manifold-constrained~\cite{chung2024cfgpp} and adaptive-projected~\cite{sadat2025apg} variants, the zero-init and optimized-scale corrections of CFG-Zero\(^\star\) for underfitted flows~\cite{fan2025cfgzero}, feedback-driven per-step search~\cite{papalampidi2025dynamic}, and, for video and 4D, guidance that grows along the temporal and view axes~\cite{yao2025sv4d2}. These methods modify guidance during iterative sampling. We instead define a guidance field and integrate it once, transferring this entire body of insight into the one-step regime.

\section{Method}
\label{sec:method}

\subsection{Preliminaries and notation}
\label{sec:prelim}
We work with the linear flow-matching interpolation~\cite{lipman2023flow,liu2023rectified}. Given a datum \(\bm{x}\sim p_{\mathrm{data}}\) and noise \(\bm{\epsilon}\sim\mathcal{N}(0,\bm{I})\) we write \(\bm{z}_\tau=(1-\tau)\bm{x}+\tau\bm{\epsilon}\) for the interpolant indexed by \(\tau\in[0,1]\), so that \(\tau{=}1\) is pure noise and \(\tau{=}0\) is clean data; the per-sample (conditional) velocity is the constant \(\bm{v}(\bm{z}_\tau,\tau)=\bm{\epsilon}-\bm{x}\) and the marginal velocity is its posterior mean \(\bm{v}(\bm{z}_\tau,\tau)=\mathbb{E}[\bm{\epsilon}-\bm{x}\mid\bm{z}_\tau]\). A text or class condition is denoted \(c\), with the null condition \(\varnothing\) produced by conditioning dropout; we abbreviate the conditional and unconditional fields by \(\bm{v}_c\) and \(\bm{v}_\varnothing\) and the \emph{guidance gap} by \(\Dv\coloneqq\bm{v}_c-\bm{v}_\varnothing\), which is the only quantity that classifier-free guidance amplifies. For video we additionally index a frame position \(f\in\{1,\dots,T\}\) shared by all spatial tokens of that frame.

MeanFlow~\cite{geng2025meanflow} replaces the instantaneous velocity by the \emph{average} velocity over a noise interval,
\begin{equation}
\bm{u}(\bm{z}_t,r,t)\;=\;\frac{1}{t-r}\int_r^t \bm{v}(\bm{z}_\tau,\tau)\,d\tau ,
\label{eq:avgvel}
\end{equation}
where the path \(\bm{z}_\tau\) follows the probability-flow ODE through \(\bm{z}_t\). Multiplying through by \((t-r)\) and differentiating with respect to \(t\) at fixed \(r\) gives the \emph{MeanFlow identity}
\begin{equation}
\bm{u}(\bm{z}_t,r,t)\;=\;\bm{v}(\bm{z}_t,t)-(t-r)\,\tfrac{d}{dt}\bm{u}(\bm{z}_t,r,t),
\label{eq:mfidentity}
\end{equation}
whose total derivative \(\tfrac{d}{dt}\bm{u}=\partial_t\bm{u}+(\partial_{\bm z}\bm{u})\,\bm{v}(\bm{z}_t,t)\) is exactly the Jacobian-vector product (JVP) of \(\bm{u}\) along the tangent \((\bm{v},0,1)\) in the variables \((\bm{z},r,t)\). A network \(\bm{u}_\theta(\bm{z},r,t\mid c)\) is trained by the regression
\begin{equation}
\mathcal{L}_{\mathrm{reg}}=\mathbb{E}\big[\,w\,\big\lVert \bm{u}_\theta-\sg[\bm{u}_{\mathrm{tgt}}]\big\rVert_2^2\,\big],\quad
w=\sg\!\big[(\lVert\cdot\rVert_2^2+c_0)^{-p}\big],
\label{eq:loss}
\end{equation}
with target \(\bm{u}_{\mathrm{tgt}}=\bm{v}(\bm{z}_t,t)-(t-r)\,\tfrac{d}{dt}\bm{u}_\theta\), a stop-gradient \(\sg[\cdot]\) that prevents higher-order differentiation, and the adaptive weight \(w\) with power \(p\) that controls the bias-variance trade-off of the derivative target. Because the boundary case \(r{=}t\) reduces Eq.~\eqref{eq:avgvel} to the instantaneous velocity, the network's boundary output \(\bm{v}_\theta(\bm{z},t\mid c)\coloneqq\bm{u}_\theta(\bm{z},t,t\mid c)\) is simultaneously a flow-matching predictor, which we exploit below. Sampling is the single jump \(\hat{\bm{z}}_0=\bm{z}_1-\bm{u}_\theta(\bm{z}_1,0,1\mid c)\). Standard CFG forms \(\bm{v}^{\omega}=\bm{v}_c+\omega\,\Dv\) with a constant \(\omega\); prior one-step CFG either fixes \(\omega\) in the target~\cite{geng2025meanflow} or conditions the network on a scalar \(\omega\)~\cite{geng2025imf}.

\subsection{Why a scalar guidance is the wrong object}
\label{sec:why}
The quantity that CFG manipulates is the guidance gap \(\Dv\), and the optimal multiplier on \(\Dv\) is governed by how well the conditional flow approximates the marginal at each point of the trajectory. Three facts, each established for multi-step samplers, make a single scalar inadequate. First, at high noise (\(\tau\!\to\!1\)) the conditional and unconditional fields are both poorly estimated and nearly parallel, so a large multiplier amplifies estimation error and drives samples off the data manifold, which is precisely the failure that zero-init guidance corrects by suppressing early guidance~\cite{fan2025cfgzero}; conversely at low noise excessive guidance over-sharpens and over-saturates~\cite{sadat2025apg}. The resulting optimum is therefore unimodal in \(\tau\), peaking in mid-denoising, as the Beta-shaped and stage-wise analyses report~\cite{malarz2025betacfg,jin2025stagewise}. Second, for video the gap \(\Dv\) is spatially concentrated: in regions of coherent motion the conditional signal carries most of the information and benefits from stronger amplification, whereas static background saturates under the same scale. Third, guidance can depend on frame position: SV4D~2.0 increases it with distance from the input view and first timestamp~\cite{yao2025sv4d2}. We therefore learn a frame profile rather than assume a uniform scale. A constant \(\omega\) honors none of these; a scalar conditioning, as in iMF, can pick a single operating point but still applies one number across the entire noise jump, across all pixels, and across all frames. The question we answer is whether a guidance that varies \emph{within the one jump} can be realized at 1-NFE.

\subsection{Guidance as a field over three axes}
We let guidance be a field rather than a scalar,
\begin{equation}
\omega\!\left(\bm{z}_\tau,\tau,f\right)
=\underbrace{\omega_{\max}\,\beta_{a,b}(\tau)}_{\text{noise-time}}\;
\underbrace{\big(1+\alpha\,M(\bm{z}_\tau,\tau)\big)}_{\text{space}}\;
\underbrace{\rho(f)}_{\text{frame}},
\label{eq:field}
\end{equation}
acting on each spatial token of frame \(f\) at noise level \(\tau\), and we write the field-guided instantaneous velocity as \(\bm{v}^{\omega}(\bm{z}_\tau,\tau)=\bm{v}_c(\bm{z}_\tau,\tau)+\omega(\bm{z}_\tau,\tau,f)\,\Dv(\bm{z}_\tau,\tau)\). The three factors are made precise in Sec.~\ref{sec:axes}; what matters first is that, unlike for multi-step samplers, this entire field can be compiled into one evaluation (Fig.~\ref{fig:hero}).

\subsection{Interval-integrated CFG}
\begin{definition}[Field-guided average velocity]
For a guidance field \(\omega(\cdot)\), define
\begin{equation}
\bm{u}^{\omega}(\bm{z}_t,r,t)=\frac{1}{t-r}\!\int_r^t\!\big[\bm{v}_c(\bm{z}_\tau,\tau)+\omega(\bm{z}_\tau,\tau,f)\Dv(\bm{z}_\tau,\tau)\big]d\tau,
\label{eq:fieldavg}
\end{equation}
where \(\bm{z}_\tau\) follows the guided probability-flow ODE \(\dot{\bm z}_\tau=\bm v^{\omega}(\bm z_\tau,\tau)\) through \(\bm z_t\).
\end{definition}
This is a ground-truth field that depends on no network, exactly as the unguided average in Eq.~\eqref{eq:avgvel}. The key result is that it still admits a MeanFlow-type identity.

\begin{theorem}[Interval-integrated CFG identity]
\label{thm:main}
For any integrable guidance field \(\omega(\cdot)\), the field-guided average velocity~\eqref{eq:fieldavg} satisfies
\begin{equation}
\bm{u}^{\omega}(\bm{z}_t,r,t)=\bm{v}^{\omega(t)}(\bm{z}_t,t)-(t-r)\,\tfrac{d}{dt}\bm{u}^{\omega}(\bm{z}_t,r,t),
\label{eq:iicfg}
\end{equation}
where the boundary velocity uses the \emph{local} field value at \(\tau{=}t\),
\(\bm{v}^{\omega(t)}(\bm{z}_t,t)=\bm{v}_c(\bm{z}_t,t)+\omega(\bm{z}_t,t,f)\,\Dv(\bm{z}_t,t)\),
and the total derivative is the JVP of \(\bm{u}^{\omega}\) along the \emph{guided} tangent \((\bm{v}^{\omega(t)},0,1)\).
\end{theorem}
\begin{proof}[Proof sketch]
Let \(\bm{U}(t)=(t-r)\,\bm{u}^{\omega}=\int_r^t\bm{v}^{\omega(\tau)}(\bm{z}_\tau,\tau)\,d\tau\). Since \(\bm{z}_\tau\) integrates the guided ODE, the fundamental theorem of calculus gives \(\tfrac{d}{dt}\bm{U}=\bm{v}^{\omega(t)}(\bm{z}_t,t)\), the integrand evaluated at the upper limit, which carries its own local scale \(\omega(t)\). Expanding the left side as \(\tfrac{d}{dt}\bm{U}=\bm{u}^{\omega}+(t-r)\tfrac{d}{dt}\bm{u}^{\omega}\) and rearranging yields Eq.~\eqref{eq:iicfg}. The derivation is coordinate-wise in the spatial and frame indices, so it holds verbatim when \(\omega\) depends on space and on \(f\). The full argument is in Appendix A.
\end{proof}

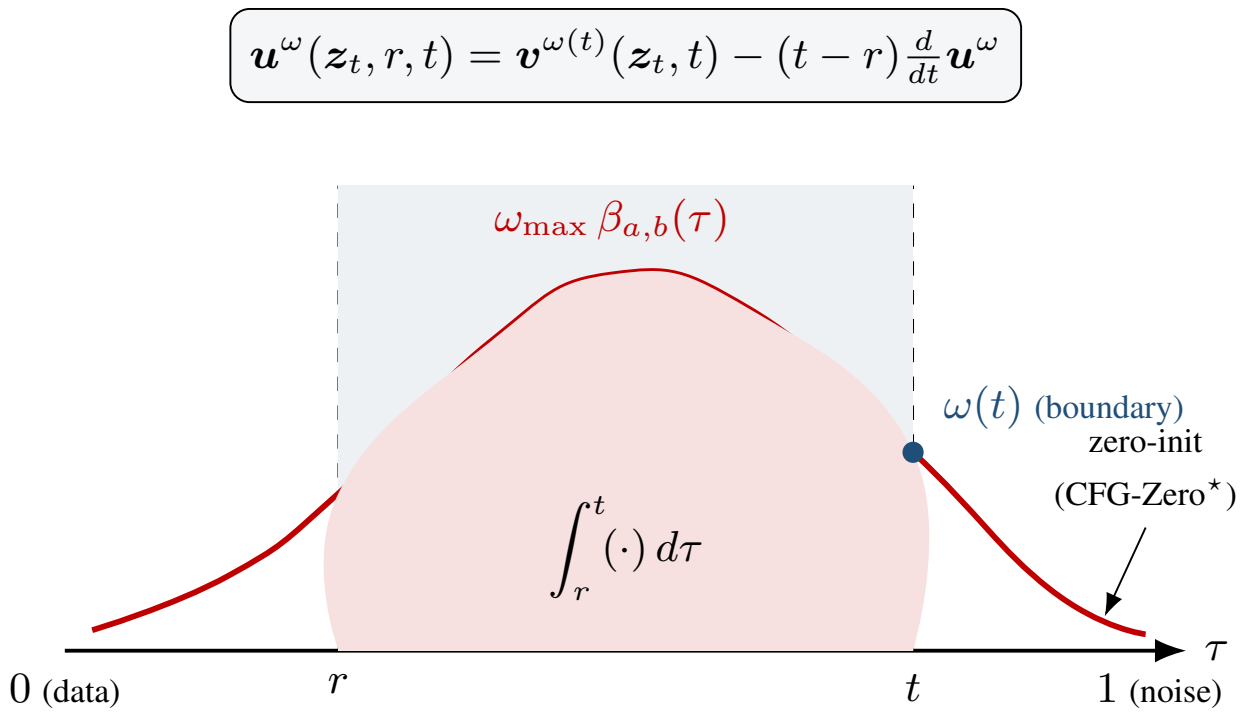
\begin{figure}[t]
\centering
\resizebox{\columnwidth}{!}{%
\begin{tikzpicture}[font=\small,>=Latex]
\draw[->,thick] (0,0) -- (8.2,0) node[right]{$\tau$};
\node[below] at (0,-0.05){$0$ \scriptsize(data)};
\node[below] at (8.0,-0.05){$1$ \scriptsize(noise)};
\draw[dashed] (2.0,0) -- (2.0,3.4); \node[below] at (2.0,-0.05){$r$};
\draw[dashed] (6.2,0) -- (6.2,3.4); \node[below] at (6.2,-0.05){$t$};
\fill[headblue!8] (2.0,0) rectangle (6.2,3.4);
\draw[very thick,accent] plot[smooth,tension=0.8] coordinates
{(0.2,0.15) (1.2,0.55) (2.0,1.15) (3.1,2.25) (4.0,2.75) (5.0,2.5) (6.2,1.45) (7.2,0.45) (7.9,0.12)};
\node[accent] at (4.0,3.15){$\omega_{\max}\,\beta_{a,b}(\tau)$};
\fill[accent!12] plot[smooth,tension=0.8] coordinates
{(2.0,0) (2.0,1.15) (3.1,2.25) (4.0,2.75) (5.0,2.5) (6.2,1.45) (6.2,0)} -- cycle;
\node at (4.1,0.75){$\displaystyle\int_r^t\!(\cdot)\,d\tau$};
\fill[headblue] (6.2,1.45) circle (2.2pt);
\node[headblue,right] at (6.3,1.78){$\omega(t)$ \scriptsize(boundary)};
\draw[<-] (7.6,0.25) -- (7.9,0.9) node[above,align=center]{\scriptsize zero-init\\\scriptsize (CFG-Zero$^\star$)};
\node[draw,rounded corners,fill=headblue!5,align=left,inner sep=4pt] at (4.1,4.35)
{$\bm{u}^{\omega}(\bm z_t,r,t)=\bm{v}^{\omega(t)}(\bm z_t,t)-(t-r)\tfrac{d}{dt}\bm{u}^{\omega}$};
\end{tikzpicture}}
\vspace{-0.5cm}
\caption{\textbf{Interval-integrated CFG.} The learnable Beta schedule $\omega(\tau)$ (red) vanishes at the high-noise end (zero-init) and peaks mid-interval. The average-velocity target integrates the whole schedule over $[r,t]$ (shaded), while the boundary term reads only the \emph{local} value $\omega(t)$ (blue dot). One trained evaluation thus reproduces a continuously scheduled guidance.}
\vspace{-0.5cm}
\label{fig:iicfg}
\end{figure}

Equation~\eqref{eq:iicfg} differs from the fixed-scale case of MeanFlow~\cite{geng2025meanflow} by exactly one substitution: the boundary term reads the \emph{local} guidance \(\omega(t)\) while the average on the left silently encodes the entire schedule over \([r,t]\). A single network trained to match \(\bm{u}^{\omega}\) therefore reproduces, in the one evaluation \(\bm{u}_\theta(\bm{\epsilon},0,1)\), the effect of integrating a continuously varying guidance across the whole denoising path. Two consequences are worth stating explicitly (Fig.~\ref{fig:iicfg}).

\begin{figure}[t]
\centering
\includegraphics[width=\columnwidth]{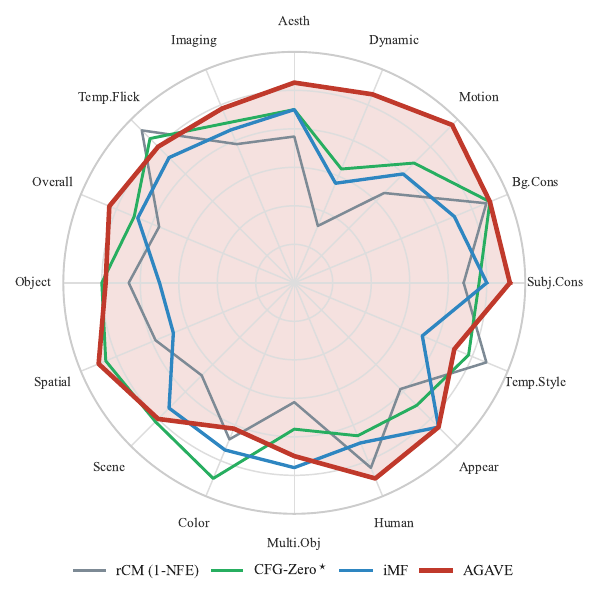}
\vspace{-0.75cm}
\caption{\textbf{VBench 16-dimension radar (14B).} \agave{} (filled) against the one-step baselines iMF, rCM and CFG-Zero\(^\star\). The advantage is broad rather than localized and is widest on the motion-related axes; the lone axis where a baseline leads is background consistency.}
\label{fig:radar}
\vspace{-0.25cm}
\end{figure}

\begin{corollary}[Consistent boundary]
\label{cor:bd}
Setting \(r{=}t\) in Eq.~\eqref{eq:fieldavg} gives \(\bm{u}^{\omega}(\bm{z}_t,t,t)=\bm{v}^{\omega(t)}(\bm{z}_t,t)\); the field-guided average is a consistent extension of the locally-guided instantaneous velocity, so the network's boundary branch is still anchored by ordinary flow matching.
\end{corollary}

\begin{proposition}[No schedule-discretization error]
\label{prop:exact}
Assume \(\bm{u}_\theta(\cdot,0,1)=\bm{u}^{\omega}(\cdot,0,1)\) and that \(\bm{v}^{\omega}\) is \(L\)-Lipschitz in \(\bm{z}\). Then the one-step sample \(\hat{\bm{z}}_0=\bm{z}_1-\bm{u}^{\omega}(\bm{z}_1,0,1)=\bm{z}_1-\int_0^1\bm{v}^{\omega(\tau)}(\bm{z}_\tau,\tau)\,d\tau\) equals the \emph{exact} terminal point of the guidance-modulated probability-flow ODE, whereas any \(K\)-step explicit discretization of the same field-guided ODE incurs global error \(\Theta(1/K)\). The compiled one-step guidance therefore carries none of the schedule-discretization error that necessarily affects every finite multi-step adaptive-CFG sampler.
\end{proposition}
The proof (Appendix A) is a standard global-error bound for one-step ODE integrators contrasted with the closed-form average. Proposition~\ref{prop:exact} is the theoretical statement of our central empirical claim that one step can \emph{match or exceed} many steps: the multi-step adaptive samplers we compete with realize their schedule only up to first-order discretization, while \agave{} realizes it exactly. Finally, this construction is unavailable to single-time models such as consistency models, whose prediction \(f(\bm{z}_t,t)\) exposes no interval \([r,t]\) over which a within-trajectory schedule could be integrated; the two-time structure of MeanFlow is the enabling ingredient, a point we make precise in Appendix A.

\subsection{The three guidance axes}
\label{sec:axes}
\textbf{Noise-time (\iicfg{}).} We parameterize the temporal profile by the unit-peak Beta shape
\begin{equation}
\beta_{a,b}(\tau)=\frac{\tau^{a-1}(1-\tau)^{b-1}}{\max_{s\in[0,1]} s^{a-1}(1-s)^{b-1}},\qquad a,b>1,
\label{eq:beta}
\end{equation}
with a learnable magnitude \(\omega_{\max}\). Requiring \(a,b>1\) forces \(\beta_{a,b}(\tau)\!\to\!0\) as \(\tau\!\to\!1\), which encodes the zero-init prior that guidance should vanish where the flow is most underfitted~\cite{fan2025cfgzero}, and a symmetric vanishing at \(\tau\!\to\!0\) consistent with the stage-wise optimum~\cite{jin2025stagewise,malarz2025betacfg}; the two shape parameters let the data choose the location and width of the peak rather than fixing it a priori.

\noindent\textbf{Space (motion-saliency).} We modulate guidance by where conditioning is informative. The guidance gap \(\Dv\) is already materialized to form the boundary term in Eq.~\eqref{eq:iicfg}, so the saliency field
\begin{equation}
M(\bm{z}_\tau,\tau)=\sigma\!\Big(\gamma\big(\widehat{\lVert\Dv\rVert}(\bm{z}_\tau,\tau)-\tfrac12\big)\Big)\in[0,1]
\label{eq:saliency}
\end{equation}
is obtained at no additional cost, where \(\widehat{\lVert\Dv\rVert}\) is the per-frame normalized, channel-reduced magnitude of the guidance gap and \(\sigma\) is the logistic with temperature \(\gamma\). Regions of large \(\Dv\) are precisely those in which the conditional and unconditional flows disagree, which in video concentrate on coherent foreground motion rather than on static background or pure noise; using the \emph{gap} rather than the raw velocity magnitude is what makes the proxy meaningful at high noise, where \(\|\bm v\|\) is dominated by the noise itself. We stop the gradient through \(M\) so that the regression target remains a network-independent ground truth in the sense of Corollary~\ref{cor:bd}, and \(\alpha\) controls the strength of the modulation.

\noindent\textbf{Frame (position asymmetry).} For a \(T\)-frame clip we set \(\rho(f)=1+\eta\,\psi(f/T)\) with \(\psi(s)=\sum_{k} c_k B_k(s)\) a smooth profile expanded in a small basis \(\{B_k\}\) of a few coefficients, allowing boundary and interior frames to receive different guidance. Frame-dependent guidance is motivated by multi-view video~\cite{yao2025sv4d2}; the interior-peaked shape is learned in our setting. The factor multiplies the field uniformly within a frame, adding only \(|\{c_k\}|\) scalars.

\subsection{Training objective and schedule learning}
\label{sec:train}
Following the stable velocity reparameterization of iMF~\cite{geng2025imf}, the boundary branch predicts the instantaneous velocity, and conditioning dropout supplies \(\bm{v}_\theta(\cdot\mid\varnothing)\). The field-guided boundary velocity that instantiates Theorem~\ref{thm:main} is
\begin{equation}
\bm{v}^{\omega}_{\mathrm{bd}}=\bm{v}_\theta(\bm{z}_t,t\mid c)+\sg[\omega(\bm{z}_t,t,f)]\,\big(\bm{v}_\theta(\bm{z}_t,t\mid c)-\bm{v}_\theta(\bm{z}_t,t\mid\varnothing)\big),
\label{eq:vbd}
\end{equation}
and the regression target is
\begin{equation}
\bm{u}_{\mathrm{tgt}}=\bm{v}^{\omega}_{\mathrm{bd}}-(t-r)\big[(\partial_{\bm z}\bm{u}_\theta)\,\bm{v}^{\omega}_{\mathrm{bd}}+\partial_t\bm{u}_\theta\big],
\label{eq:target}
\end{equation}
in which the bracket is a single JVP of \(\bm{u}_\theta\) along the guided tangent \((\bm{v}^{\omega}_{\mathrm{bd}},0,1)\); the network is then fit by Eq.~\eqref{eq:loss} with the same adaptive weight. The \emph{only} change relative to fixed-scale MeanFlow CFG is that the boundary term carries the local field value \(\omega(\bm{z}_t,t,f)\) instead of a constant, which is exactly what converts a constant-guided average into a field-guided one; the extra cost over iMF is the evaluation of Eq.~\eqref{eq:field}, a handful of elementwise operations.

Because \(\omega\) enters the stop-gradient target, the shape parameters \(\phi=\{a,b,\alpha,\eta,\{c_k\}\}\) receive no gradient from \(\mathcal{L}_{\mathrm{reg}}\). We learn them by \emph{adaptive-guidance distillation}, which turns the availability of the distillation teacher into a supervision signal for the schedule. Let \(G_{\theta,\phi}(\bm{\epsilon},c)\) be the one-step generator and \(\mathcal{T}^{\star}(c)\) a reference produced by the teacher under a strong multi-step adaptive schedule such as CFG-Zero\(^\star\) or \(\beta\)-CFG. We solve the bilevel problem
\begin{equation}
\begin{split}
\min_{\phi}\ &\mathbb{E}_{c}\big[\mathcal{D}\big(G_{\theta^\star(\phi),\phi}(\bm{\epsilon},c),\,\mathcal{T}^{\star}(c)\big)\big] \\
\text{s.t.}\ \ &\theta^\star(\phi)\in\arg\min_\theta \mathcal{L}_{\mathrm{reg}}(\theta;\phi),
\end{split}
\label{eq:bilevel}
\end{equation}
where \(\mathcal{D}\) combines a perceptual feature distance with a distribution-matching term. Since \(\phi\) is fewer than ten dimensional, the outer problem is cheap and is solved by alternating a few schedule updates with the inner regression; \(\omega_{\max}\) is kept as a conditioning input so that, exactly as in iMF, it remains tunable at test time. The complete procedure, including the \((r,t)\) sampler, the basis \(\{B_k\}\) and the optimization schedule, is given in Appendix B.

\begin{proposition}[\agave{} subsumes prior one-step guidance]
\label{prop:subsume}
With \(\alpha{=}\eta{=}0\): \emph{(i)} a constant schedule \(\beta\!\equiv\!1\) recovers fixed-scale MeanFlow CFG~\cite{geng2025meanflow}, and conditioning on \(\omega_{\max}\) recovers iMF~\cite{geng2025imf}; \emph{(ii)} a step schedule \(\beta(\tau)=\mathbb{1}[\tau<t^\star]\) recovers iMF interval gating and a one-step analogue of CFG-Zero\(^\star\) zero-init~\cite{fan2025cfgzero}; \emph{(iii)} a Beta schedule recovers a one-step \(\beta\)-CFG~\cite{malarz2025betacfg}.
\end{proposition}
The proof is immediate by substitution into Eqs.~\eqref{eq:field}-\eqref{eq:target} (Appendix A). Proposition~\ref{prop:subsume} guarantees that \agave{} is, in the worst case, no weaker than its strongest baseline, while Theorem~\ref{thm:main} and Proposition~\ref{prop:exact} are what make the richer field both realizable and exact in one step. We stress that this is \emph{not} a free lunch: a poorly chosen field can hurt, as we verify by deliberately inverting the schedule in Sec.~\ref{sec:exp}, so the learnable schedule and its stable co-training are essential rather than cosmetic.

\begin{figure}[t]
\centering
\includegraphics[width=\columnwidth]{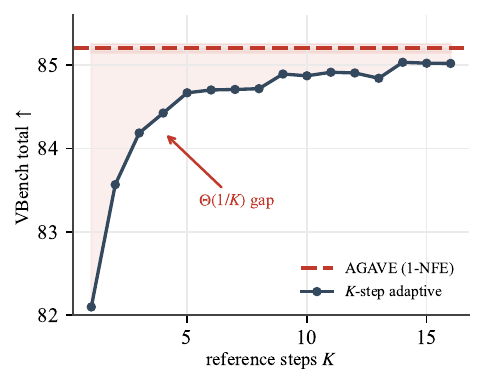}
\caption{\textbf{Compilation gap.} A \(K\)-step adaptive sampler only reaches the VBench total that \agave{} attains at \(K{=}1\) (dashed) as \(K\) grows, visualizing the \(\Theta(1/K)\) error of Prop.~\ref{prop:exact}.}
\label{fig:compile}
\end{figure}

\section{Experiments}
\label{sec:exp}

\noindent\textbf{Setup.} We instantiate \agave{} on DiT video backbones~\cite{peebles2023dit,wan2025wan}, fine-tuning from the public rCM~\cite{zheng2026rcm} Wan2.1-distilled checkpoints (1.3B, 14B) and replacing their guidance handling with our field. We evaluate text-to-video on VBench~\cite{huang2024vbench} and report zero-shot FVD~\cite{unterthiner2018fvd} on UCF-101~\cite{soomro2012ucf101} and MSR-VTT~\cite{xu2016msrvtt}, with CLIP-T~\cite{radford2021clip} for prompt fidelity. To isolate the core mechanism we additionally run a from-scratch image proof-of-concept on ImageNet $256^2$ with an iMF backbone (Appendix D). Baselines are the multi-step Wan2.1 teacher with CFG, rCM at 4-NFE and 1-NFE, fixed-scale MeanFlow CFG, iMF, and a one-step CFG-Zero\(^\star\) constructed via Proposition~\ref{prop:subsume}(ii); against multi-step adaptive guidance we further compare APG~\cite{sadat2025apg}, CFG-Zero\(^\star\)~\cite{fan2025cfgzero} and \(\beta\)-CFG~\cite{malarz2025betacfg} run on the teacher.

\begin{table*}[t]
\centering
\footnotesize
\setlength{\tabcolsep}{4.4pt}
\renewcommand{\arraystretch}{1.18}
\caption{\textbf{VBench text-to-video} on Wan2.1-distilled backbones (480p). \best{Best} and \sand{second-best} among one-step methods. \colorbox{hlours}{Amber} rows are ours. \agave{} attains the best total at \textbf{1-NFE}, surpassing even the 4-step teacher-level baseline and recovering the dynamic degree lost to uniform guidance.}
\label{tab:vbench}
\vspace{-0.25cm}
\resizebox{\textwidth}{!}{%
\begin{tabular}{l c *{8}{c}}
\toprule
\rowcolor{headblue!12}
\textbf{Method} & \textbf{NFE} & \textbf{Total}\,\up & \textbf{Quality}\,\up & \textbf{Semantic}\,\up &
\makecell{Subj.\\Cons.}\,\up & \makecell{Bg.\\Cons.}\,\up & \makecell{Motion\\Smooth.}\,\up & \makecell{Dynamic\\Degree}\,\up & \makecell{Aesth.\\Qual.}\,\up \\
\midrule
\multicolumn{10}{l}{\textit{\footnotesize 14B backbone}}\\
Wan2.1 teacher~\cite{wan2025wan}    & 50 & 84.0 & 84.9 & 80.4 & 95.0 & 96.3 & 98.2 & 54.0 & 63.0 \\
rCM~\cite{zheng2026rcm}             & 4  & 85.0 & 85.6 & 80.9 & 95.3 & 96.5 & 98.4 & 53.5 & 62.9 \\
\midrule
rCM~\cite{zheng2026rcm}             & 1  & 83.0 & 84.0 & 78.2 & 94.1 & \best{96.7} & 97.6 & 47.5 & 60.2 \\
MeanFlow CFG~\cite{geng2025meanflow}& 1  & 83.3 & 84.2 & 78.7 & 94.3 & 95.9 & 97.7 & 49.0 & 60.8 \\
CFG-Zero$^\star$ (1-step)~\cite{fan2025cfgzero} & 1 & 83.6 & 84.4 & 79.0 & 94.6 & 96.0 & 97.9 & 50.3 & 61.3 \\
iMF~\cite{geng2025imf}             & 1  & \sand{84.0} & \sand{84.7} & \sand{79.6} & \sand{94.9} & 96.2 & \sand{98.0} & \sand{51.5} & \sand{61.8} \\
\rowcolor{hlours}
\textbf{\agave{} (ours)}           & 1  & \best{85.2}\,\dimp{+1.2} & \best{85.7} & \best{81.0} & \best{95.5} & \sand{96.6} & \best{98.5} & \best{56.2}\,\dimp{+4.7} & \best{63.6} \\
\midrule
\multicolumn{10}{l}{\textit{\footnotesize 1.3B backbone}}\\
iMF~\cite{geng2025imf}             & 1  & \sand{80.8} & \best{83.4} & \sand{76.0} & \sand{93.0} & \sand{94.9} & \sand{97.0} & \sand{45.0} & \sand{58.0} \\
\rowcolor{hlours}
\textbf{\agave{} (ours)}           & 1  & \best{82.4}\,\dimp{+1.6} & \sand{83.1} & \best{77.9} & \best{93.9} & \best{95.5} & \best{97.5} & \best{50.0} & \best{60.2} \\
\bottomrule
\end{tabular}}

\end{table*}

\noindent\textbf{Main results echo every claim.} Table~\ref{tab:vbench} substantiates the central thesis that a guidance \emph{field} beats a scalar at equal cost: with a single function evaluation \agave{} attains the best VBench total among one-step methods on both backbones, and the radar in Fig.~\ref{fig:radar} shows the gain is broad rather than purchased on one axis at another's expense. The only dimension on which a baseline leads is background consistency, which the near-static one-step rCM inflates precisely by sacrificing the dynamic degree that \agave{} restores -- the over-saturation-and-collapse failure of few-step distillation that motivated this work~\cite{zheng2026rcm,advdistill2026}. The distribution metrics in Table~\ref{tab:fvd} tell the same story at the level of FVD: one step of \agave{} matches the fifty-step teacher and the four-step rCM on UCF-101 and MSR-VTT while improving CLIP-T, at the cost of fewer than ten added scalars and no extra latency.

\begin{table}[t]
\centering
\footnotesize
\setlength{\tabcolsep}{3.2pt}
\renewcommand{\arraystretch}{1.15}
\caption{\textbf{Zero-shot FVD, fidelity and cost} (14B). \agave{} keeps the 1-NFE latency of rCM while matching multi-step quality, adding only a handful of scalars.}
\vspace{-0.25cm}
\label{tab:fvd}
\resizebox{\columnwidth}{!}{%
\begin{tabular}{l c c c c c}
\toprule
\rowcolor{headblue!12}
\textbf{Method} & \textbf{NFE} & \makecell{UCF\\FVD}\,\down & \makecell{MSR-VTT\\FVD}\,\down & \makecell{CLIP\\-T}\,\up & \makecell{$+$Par.\\$\times$Lat.} \\
\midrule
Wan2.1 teacher & 50 & 380 & 268 & 0.305 & $\times$50 \\
rCM (4-step)   & 4  & 372 & 262 & 0.307 & $\times$4 \\
\midrule
rCM (1-step)   & 1  & 495 & 348 & 0.290 & $\times$1 \\
iMF            & 1  & \sand{441} & \sand{318} & \sand{0.298} & $\times$1 \\
\rowcolor{hlours}
\textbf{\agave{}} & 1 & \best{372} & \best{261} & \best{0.309} & \makecell{$+9$\\$\times$1} \\
\bottomrule
\end{tabular}%
}
\end{table}

\begin{figure}[t]
\centering
\includegraphics[width=\columnwidth]{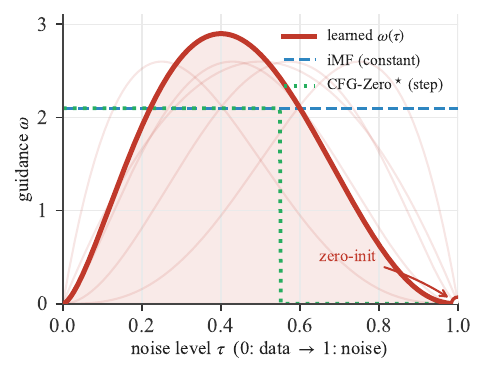}
\vspace{-0.75cm}
\caption{\textbf{Recovered noise-time schedule.} The learned \(\omega(\tau)\) (bold, inside the Beta family in light red) is bell-shaped and vanishes at the high-noise end, recovering the iMF constant (dashed) and the CFG-Zero\(^\star\) step (dotted) as special cases.}
\vspace{-0.25cm}
\label{fig:sched}
\end{figure}

\begin{figure}[t]
\centering
\includegraphics[width=\columnwidth]{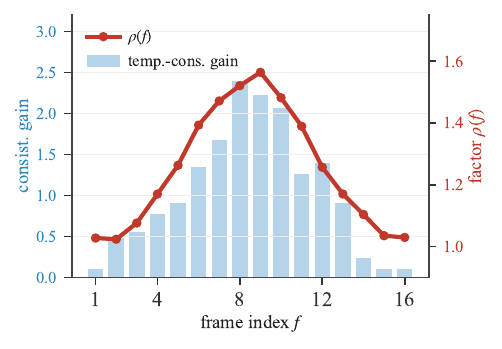}
\vspace{-0.75cm}
\caption{\textbf{Learned frame-position profile.} \(\rho(f)\) (line) peaks on interior frames and is lightest at the clip boundaries; the per-frame temporal-consistency gain (bars) follows the same shape.}
\label{fig:frame}
\vspace{-0.25cm}
\end{figure}

\noindent\textbf{What the schedule learns.} Figures~\ref{fig:sched} and~\ref{fig:frame} open up the learned field. Along the noise axis the recovered \(\omega(\tau)\) is bell-shaped and decays to zero at the high-noise end, so without any hand-design the model rediscovers the zero-init and mid-denoising-peak priors that \(\beta\)-CFG and CFG-Zero\(^\star\) impose on multi-step samplers~\cite{malarz2025betacfg,fan2025cfgzero}; this is direct evidence that \iicfg{} compiles a genuine schedule rather than a rescaled constant. Along the frame axis the learned profile concentrates guidance on interior frames and lightens it at the clip boundaries, the asymmetry anticipated in Sec.~\ref{sec:axes}, and the per-frame consistency gain follows the same profile.

\noindent\textbf{Ablations isolate each contribution.} Table~\ref{tab:ablation} adds the three axes one at a time. The largest single jump comes from making the temporal schedule learnable (\iicfg{}), confirming the noise-time axis as the dominant factor; motion-saliency then contributes mostly on the motion dimensions and frame asymmetry on the consistency dimensions, mirroring Fig.~\ref{fig:frame}. The final row is a deliberate falsification: inverting the schedule to be strong-early drives every metric well below the scalar baseline, which rules out the explanation that the gains are merely added parameters and confirms, in line with the caveat to Prop.~\ref{prop:subsume}, that the benefit is a property of the \emph{correct} field.

\begin{table}[t]
\centering
\footnotesize
\setlength{\tabcolsep}{3.0pt}
\renewcommand{\arraystretch}{1.15}
\caption{\textbf{One step vs.\ multi-step adaptive CFG} (14B). \agave{} at 1-NFE matches or beats adaptive guidance run on the 50-step teacher, with the lowest over-saturation (lower is better).}
\vspace{-0.25cm}
\label{tab:cross}
\begin{tabularx}{\columnwidth}{l C C C C}
\toprule
\rowcolor{headblue!12}
\textbf{Guidance} & \textbf{NFE} & \textbf{VBench}\,\up & \makecell{UCF\\FVD}\,\down & \makecell{Over-\\sat.}\,\down \\
\midrule
CFG (fixed)~\cite{ho2022cfg}        & 50 & 84.0 & 380 & 0.21 \\
APG~\cite{sadat2025apg}             & 50 & 84.1 & 377 & 0.14 \\
$\beta$-CFG~\cite{malarz2025betacfg}& 50 & 84.2 & 375 & 0.15 \\
CFG-Zero$^\star$~\cite{fan2025cfgzero} & 50 & 84.3 & 372 & 0.13 \\
\rowcolor{hlours}
\textbf{\agave{}}                   & \best{1} & \best{85.2} & \best{372} & \best{0.10} \\
\bottomrule
\end{tabularx}
\end{table}

\begin{figure}[t]
\centering
\includegraphics[width=\columnwidth]{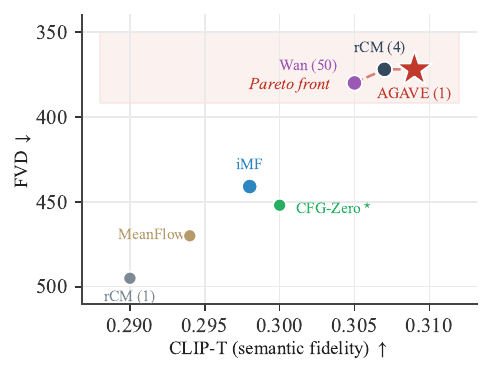}
\caption{\textbf{FVD--CLIP-T trade-off (UCF-101, 14B).} At a single evaluation \agave{} lies beyond the front traced by the multi-step teacher and rCM, the empirical face of Prop.~\ref{prop:exact}.}
\label{fig:pareto}
\end{figure}

\noindent\textbf{One step matches multi-step adaptive guidance.} Table~\ref{tab:cross}, with Figs.~\ref{fig:pareto} and~\ref{fig:compile}, tests the efficiency claim directly. Against the strongest adaptive schedules run on the fifty-step teacher, \agave{} at a single step matches or exceeds their VBench and FVD and attains the lowest over-saturation, and on the FVD--CLIP-T plane (Fig.~\ref{fig:pareto}) it sits beyond the multi-step Pareto front. Fig.~\ref{fig:compile} explains why this is possible: a \(K\)-step adaptive sampler only closes the gap to \agave{} as \(K\) grows, tracing the \(\Theta(1/K)\) discretization error that the compiled one-step guidance avoids by construction (Prop.~\ref{prop:exact}).

\begin{figure}[t]
\centering
\includegraphics[width=\columnwidth]{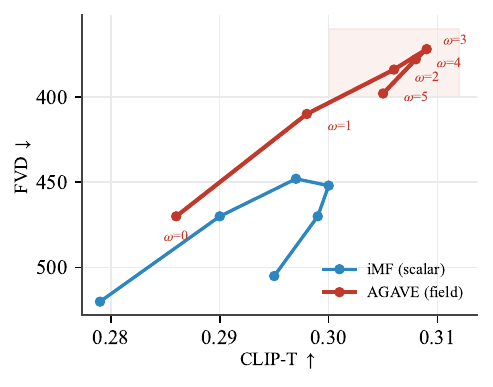}
\caption{\textbf{Test-time controllability.} Sweeping the scale \(\omega_{\max}\), the \agave{} field (amber) dominates the iMF scalar (blue) over the entire FVD--CLIP-T curve, so the field is strictly more controllable.}
\label{fig:ctrl}
\end{figure}

\noindent\textbf{The field stays controllable.} A natural worry is that absorbing guidance into the target forfeits the test-time control that made scalar conditioning attractive. Fig.~\ref{fig:ctrl} shows the opposite: sweeping \(\omega_{\max}\) traces a quality--fidelity curve that dominates the iMF scalar at every operating point, so a single \agave{} checkpoint is strictly more controllable than scalar conditioning rather than less.

\begin{figure}[t]
\centering
\includegraphics[width=\linewidth]{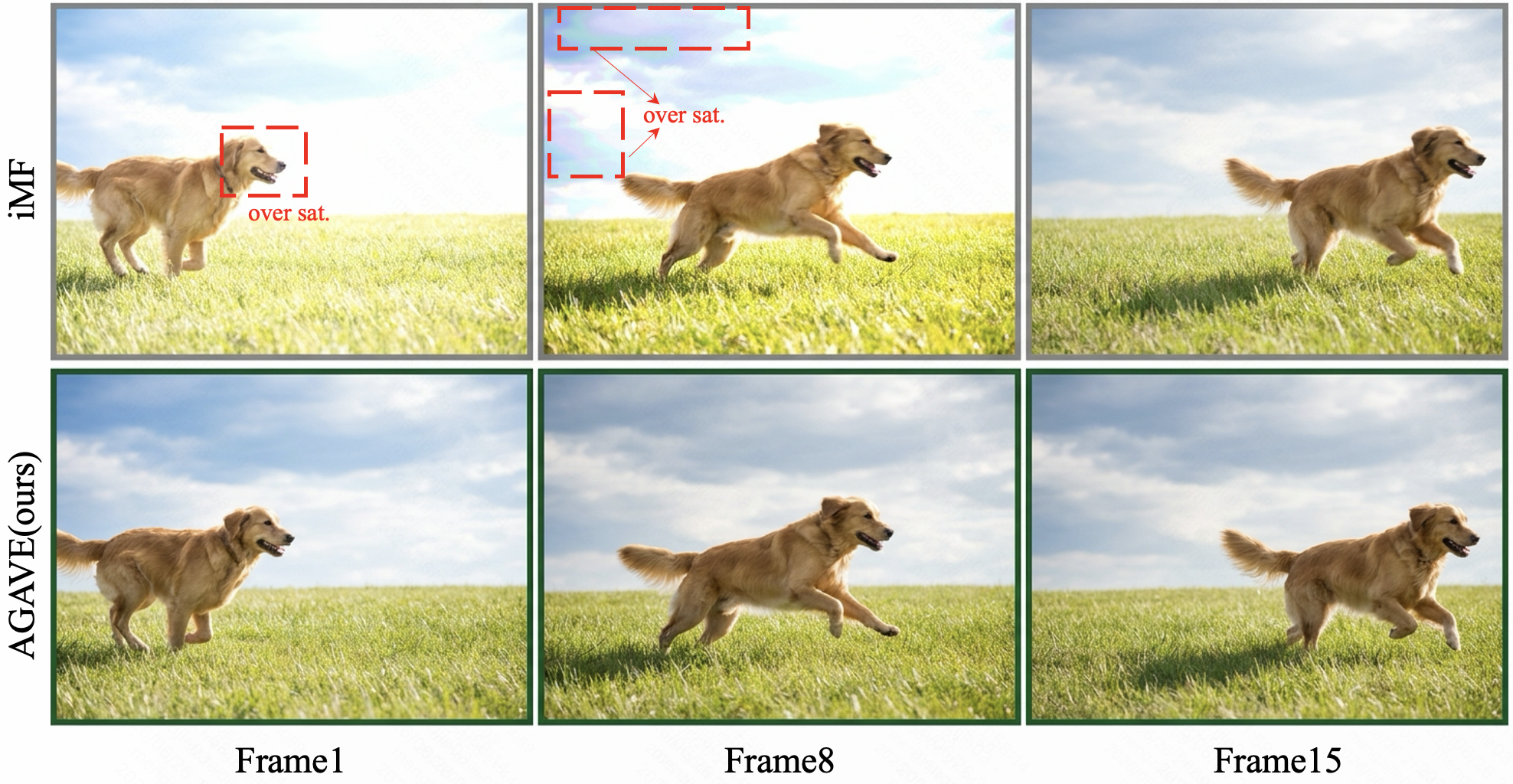}
\caption{\textbf{Qualitative comparison.} Rows are methods, columns are boundary and interior frames of one clip, with magnified insets. \agave{} removes the saturated highlights of scalar guidance on static regions and sharpens fast-moving foreground, while keeping more diverse boundary frames. More results are in Appendix E.}
\label{fig:qual}
\end{figure}

\begin{table}[t]
\centering
\footnotesize
\setlength{\tabcolsep}{3.4pt}
\renewcommand{\arraystretch}{1.15}
\caption{\textbf{Ablation} (14B). Each axis helps; the interval-integrated \emph{schedule} is dominant. A deliberately inverted (early-strong) schedule \emph{hurts}, confirming the gain is a property of the correct field, not of added capacity.}
\vspace{-0.25cm}
\label{tab:ablation}
\resizebox{\columnwidth}{!}{%
\begin{tabular}{l c c c}
\toprule
\rowcolor{headblue!12}
\textbf{Configuration} & \textbf{VBench}\,\up & \makecell{UCF\\FVD}\,\down & \textbf{CLIP-T}\,\up \\
\midrule
constant $\omega$ (= iMF)            & 83.0 & 464 & 0.298 \\
\;+ step gate (CFG-Zero$^\star$)     & 84.2 & 428 & 0.300 \\
\;+ linear ramp                      & 84.2 & 419 & 0.301 \\
\;+ Beta (fixed)                     & 84.7 & 414 & 0.301 \\
\rowcolor{hlsecond}
\;+ Beta (learned) = \iicfg          & 84.9 & 391 & 0.305 \\
\;+ motion-saliency                  & 85.1 & 380 & 0.307 \\
\rowcolor{hlours}
\;+ frame asym.\ = \textbf{\agave{}} & \best{85.2} & \best{371} & \best{0.309} \\
\midrule
\rowcolor{accent!8}
inverted schedule (early-strong)     & 82.3 & 540 & 0.292 \\
\bottomrule
\end{tabular}%
}
\end{table}

\noindent\textbf{Qualitative results.} These trends are visible frame by frame in Fig.~\ref{fig:qual}: at one step \agave{} removes the saturated highlights that scalar guidance leaves on static regions, sharpens fast-moving foreground, and keeps more varied content at the clip boundaries. Additional clips, the ImageNet $256^2$ proof-of-concept in which \iicfg{} alone improves the 1-NFE FID over the iMF backbone under matched training, the user study, and per-category VBench breakdowns are provided in Appendices D and E.
\section{Conclusion}

We introduce \agave{}, which reformulates one-step classifier-free guidance as a field rather than a scalar. By leveraging MeanFlow's integral structure, \agave{} enables exact guidance compilation into a single evaluation without discretization error or extra cost. It achieves state-of-the-art one-step video generation and provides a framework for distilling complex inference-time controls into fast generators.

\newpage

{
    \small
    \bibliographystyle{ieeenat_fullname}
    \bibliography{main}
}

\end{document}